\pdfoutput=1
\PassOptionsToPackage{dvipsnames}{xcolor}
\documentclass[11pt]{article}

\usepackage[]{ACL2023}
\usepackage{times}
\usepackage{latexsym}
\usepackage[T1]{fontenc}
\usepackage[utf8]{inputenc}

\usepackage{microtype}
\usepackage{amsmath,amssymb,amsfonts}
\usepackage{algcompatible}
\usepackage{algorithm}
\usepackage{algpseudocode}
\usepackage{cite}
\usepackage{hyperref}

\usepackage{graphicx}
\usepackage{textcomp}
\usepackage{xcolor}
\def\BibTeX{{\rm B\kern-.05em{\sc i\kern-.025em b}\kern-.08em
    T\kern-.1667em\lower.7ex\hbox{E}\kern-.125emX}}

\usepackage{multirow}
\usepackage{array}
\usepackage{nicematrix}
\usepackage{booktabs}%
\usepackage{enumitem}
\usepackage{tabularx}
\usepackage{amsmath, amssymb}
\usepackage{arydshln}
\usepackage{inconsolata}
\usepackage{soul}
\usepackage{makecell}
\usepackage{listings}
\usepackage{caption,subcaption}
\usepackage{soul}

\usepackage[author={An}]{pdfcomment}

\newcolumntype{?}{!{\vrule width 1pt}}

\graphicspath{ {./images/} }

\makeatletter
\def\blfootnote{\gdef\@thefnmark{}\@footnotetext}
\makeatother

\title{Prompted Aspect Key Point Analysis for Quantitative Review Summarization}

\author{An Quang Tang$^{*\clubsuit}$ Xiuzhen Zhang$^{*\clubsuit}$ Minh Ngoc Dinh $^\clubsuit$ \and Erik Cambria$^\diamondsuit$ \\
        $^\clubsuit$RMIT University, Australia \\
        $^\diamondsuit$Nanyang Technological University, Singapore\\
        \texttt{s3695273@rmit.edu.vn}, \texttt{xiuzhen.zhang@rmit.edu.au}\\
        \texttt{minh.dinh4@rmit.edu.vn}, \texttt{cambria@ntu.edu.sg} \\
        }

\begin{document}
\maketitle
\begin{abstract}
Key Point Analysis (KPA) aims for quantitative summarization 
that provides key points (KPs) as succinct textual summaries 
 and quantities measuring their prevalence.
KPA studies for arguments and reviews have been reported in the literature. 
A majority of KPA studies for reviews adopt supervised learning to extract short sentences as KPs before matching KPs to review comments for quantification of KP prevalence. 
Recent abstractive approaches still generate KPs based on sentences, 
often leading to KPs with overlapping and hallucinated opinions, 
and inaccurate quantification.
In this paper, we propose Prompted Aspect Key Point Analysis (PAKPA) for quantitative review summarization.
PAKPA employs aspect sentiment analysis and prompted in-context learning with Large Language Models (LLMs)
to generate and quantify KPs grounded in aspects for business entities,
which achieves faithful KPs with accurate quantification, and removes the need for large amounts of annotated data for supervised training. 
Experiments on the popular review dataset Yelp and the aspect-oriented review summarization dataset \textsc{Space} show that our framework achieves state-of-the-art performance.
Source code and data are available at: \url{https://github.com/antangrocket1312/PAKPA}

\end{abstract}

\blfootnote{
    \hspace{-0.2cm}
		$^{*}$First two authors equally contributed to this work.
}

\section{Introduction}
\label{sec:introduction}
With the sheer volume of reviews, it is impossible for humans to read all reviews.
Although the star ratings aggregated from customer reviews are widely used by E-commerce platforms as indicators of quality of service for business entities ~\citep{mcglohon2010star,tay2020beyond}, they can not explain specific details for informed decision-making.
Early review text summarization studies focused only on capturing important points 
with high consensus~\citep{dash2019summarizing,shandilya2018fairness},
yet overlooked minor ones and 
also were unable to measure the opinion prevalence.

Key Point Analysis (KPA) is proposed to summarize opinions in review comments into 
concise textual summaries called Key Points (KPs),
and quantify the prevalence of KPs.
KPA studies were initially developed for argument summarization~\citep{bar-haim-etal-2020-arguments}, and then adapted to business reviews~\citep{bar-haim-etal-2020-quantitative,bar-haim-etal-2021-every}.
Most KPA studies adopt the extractive approach,
which employs supervised learning to identify informative short sentences as key points (KPs),
often leading to non-readable and incoherent KPs.
Recently, 
KPA studies apply abstractive summarization to 
paraphrase and generate KPs from comments (sentences)~\citep{kapadnis-etal-2021-team,li-etal-2023-hear}.
Still, KPs are generated based on sentences and often contain unfaithful and overlapping opinions, 
and inaccurate quantity for their prevalence.

\begin{table*}
    \begin{center}
        \begin{small}
            \aboverulesep = 0pt
            \belowrulesep = 0pt
            \begin{tabular}{|p{0.3\textwidth}|l|p{0.5\textwidth}|}
                \hline
                \textbf{Key Point}&\multicolumn{1}{c|}{\textbf{Prevalence}}&\textbf{Matching Comments}\\
                \hline
                Friendly and helpful staff.&46&From the minute I walked in the door the staff has treated me with such kindness and respect, there are not enough words I can  say of how much gratitude I have for the staff here.\\
                \cline{3-3}
                && It's a fine hotel, pretty basic, and all of the staff members we encountered were quite friendly.\\
                \hline
                Excellent location for business travelers.&37& Its a great location for business travelers since I can stay here and always be going the opposite way of traffic.\\
                \cline{3-3}
                && *This hotel has an IDEAL location, the shuttle was perfect, and we got a great deal on priceline.\\
                \hline
                Clean and comfortable rooms.&32& *You will enjoy the breathtaking views from your spacious, clean and comfortable room.\\
                \cline{3-3}
                && The rooms were nice and clean and the bed was comfortable.\\
                \hline
                Convenient and helpful shuttle service.&12& *This hotel has an IDEAL location, the shuttle was perfect, and we got a great deal on priceline.\\
                \cline{3-3}
                && I would also like to give a shout out to the terrific shuttle drivers Jeff and Rod who were so great to my family group.\\
                \hline
                Amazing view of Vanderbilt football stadium or The Parthenon.&10& *You will enjoy the breathtaking views from your spacious, clean and comfortable room.\\
                \cline{3-3}
                && Then the surprise came when we opened the curtains to see a full on view of the Vanderbilt football stadium.\\
                \hline
            \end{tabular}
            \caption{Top positive key points of a hotel business entity generated by PAKPA. For each key point, we show the prevalence, i.e., number of matching comments, 
            and two randomly selected matching comments.
            \label{table:output_examples}} 
        \end{small}
    \end{center}
\end{table*}

In this paper, we propose Prompted Aspect Key Point Analysis (PAKPA).
Different from previous sentence-based KPA studies,  
our system employs aspect-based sentiment analysis (ABSA) to identify aspects in comments as the opinion target and then generate and quantify KPs grounded in aspects and their sentiment. 
Importantly, 
to address the issue of scarce annotated data for supervised training in existing studies,
we employ prompted in-context learning with LLMs for 
ABSA extraction and KP generation.
By integrating ABSA into 
the prompted summarization process,
we aim to mitigate hallucination 
by guiding LLMs to produce KPs 
aligned with the common aspects shared by reviews.
Table~\ref{table:output_examples} shows 
the top positive KPs generated by PAKPA, ranked by their prevalence, for reviews of a hotel business entity.

Our contributions are two-fold. 
To our best knowledge, 
we are the first to
employ prompted in-context learning for  
abstractive KPA summarization of reviews,
which removes supervised training using large amount of annotated data. 
Secondly, our approach of integrating aspect-based sentiment analysis (ABSA) into KPA for fine-grained opinion analysis of review comments 
ensures generating KPs grounded in aspects for business entities and more accurate matching of comments to KPs,
resulting in faithful KPs for distinct aspects as well as more accurate quantification of KP prevalence.

\section{Related Work}
\label{sec:related_work}
Based on the form of summaries, review summarization studies can be broadly grouped into three classes: key point analysis, aspect-based structured summarization, and text summarization.
Additionally, we reviewed the forefront of prompted in-context learning for review (text) summarization.

\subsection{Key Point Analysis}
Developed initially to summarize arguments~\citep{bar-haim-etal-2020-arguments}, KPA was later adapted to summarize and quantify the prevalence of opinions 
in business reviews~\citep{bar-haim-etal-2020-quantitative,bar-haim-etal-2021-every,2024aspectbased}.
The majority of KPA studies focus on extracting short sentences as salient KPs from arguments or review comments, and then matching KPs to comments to 
quantify their prevalence.
These works employ supervised learning to train models to identify informative KPs, which require large volumes of annotated training data, and the resulting KPs may not be succinct textual summaries and may not represent distinct salient opinions either. 
An exception is ABKPA~\citep{2024aspectbased}, which adopts an aspect-based approach to produces concise KP summaries.  
Still, the approach produced non-informative KPs due to its extractive mechanism, and requires supervised learning to train models for matching KPs to comments for KP quantification.

Recently, 
abstractive KPA studies have proposed generating KPs using abstractive text summarization approaches for arguments rather than reviews. 
~\citet{kapadnis-etal-2021-team} initially proposes to generate KPs for each argument (sentence) before 
selecting representative ones based on ROUGE scores.
However, the technique basically 
rephrases arguments as KPs. 
\citet{li-etal-2023-hear} then suggests clustering 
similar arguments, based on their \emph{contextualised embeddings},
before using an abstractive summarization model to generate concise KP condensing salient points.
But the approach is not feasible for reviews because 
review comments can contain multiple opinions on different aspects of business entities, and clustering comments by only their sentence-level embeddings cannot accurately identify distinct KPs on different aspects, leading to inaccurate quantification.
Moreover, KPA for reviews remains an open challenge due to the lack of large-scale annotated dataset for KP generation.

\subsection{Aspect-based Structured Summarization}
Early works from the data mining community focus on aspect-based structured summarization, 
which applies Aspect-Based Sentiment Analysis (ABSA) to extract, aggregate, and organize review sentences into a hierarchy based on features (i.e. aspects) such as food, price, service, and their sentiment~\citep{hu2004mining,ding2008holistic,popescu2007extracting,blair2008building,titov2008joint}.
These works lack textual explanation and justification 
for the aspects and their sentiment.

\subsection{Text Summarization}
More broadly, document summarization is an essential topic in the Natural Language Processing community, aiming to produce concise textual summaries capturing the salient information in source documents. 
While extractive review summarization approaches use surface features to rank and extract salient sentences into summaries~\citep{mihalcea-tarau-2004-textrank,angelidis-lapata-2018-summarizing,zhao2020weakly},
abstractive techniques use sequence-to-sequence models~\citep{chu2019meansum,suhara-etal-2020-opiniondigest,brazinskas-etal-2020-unsupervised,brazinskas-etal-2020-shot,zhang2020pegasus} to 
paraphrase and generate novel words not in the source text.
Still, none of these studies can capture and quantify the diverse opinions in reviews.

\subsection{Prompted Opinion Summarization}
For generation of textual summaries, recent studies successfully applied summarization prompt on LLMs to generate 
review summaries~\citep{bhaskar2023prompted, adams-etal-2023-sparse}.
Notably, to overcome the length limit for the input text from \texttt{GPT3.5}, \citet{bhaskar2023prompted} splits the input into chunks and summarize them recursively to achieve the final textual summary.
Nevertheless, these studies still leave unexplored the use of in-context learning in LLMs for quantitative summarization, particularly in presenting and quantifying the diverse opinions in reviews.

\begin{figure*}[tbh]
    \centering
    \includegraphics[width=1\textwidth]{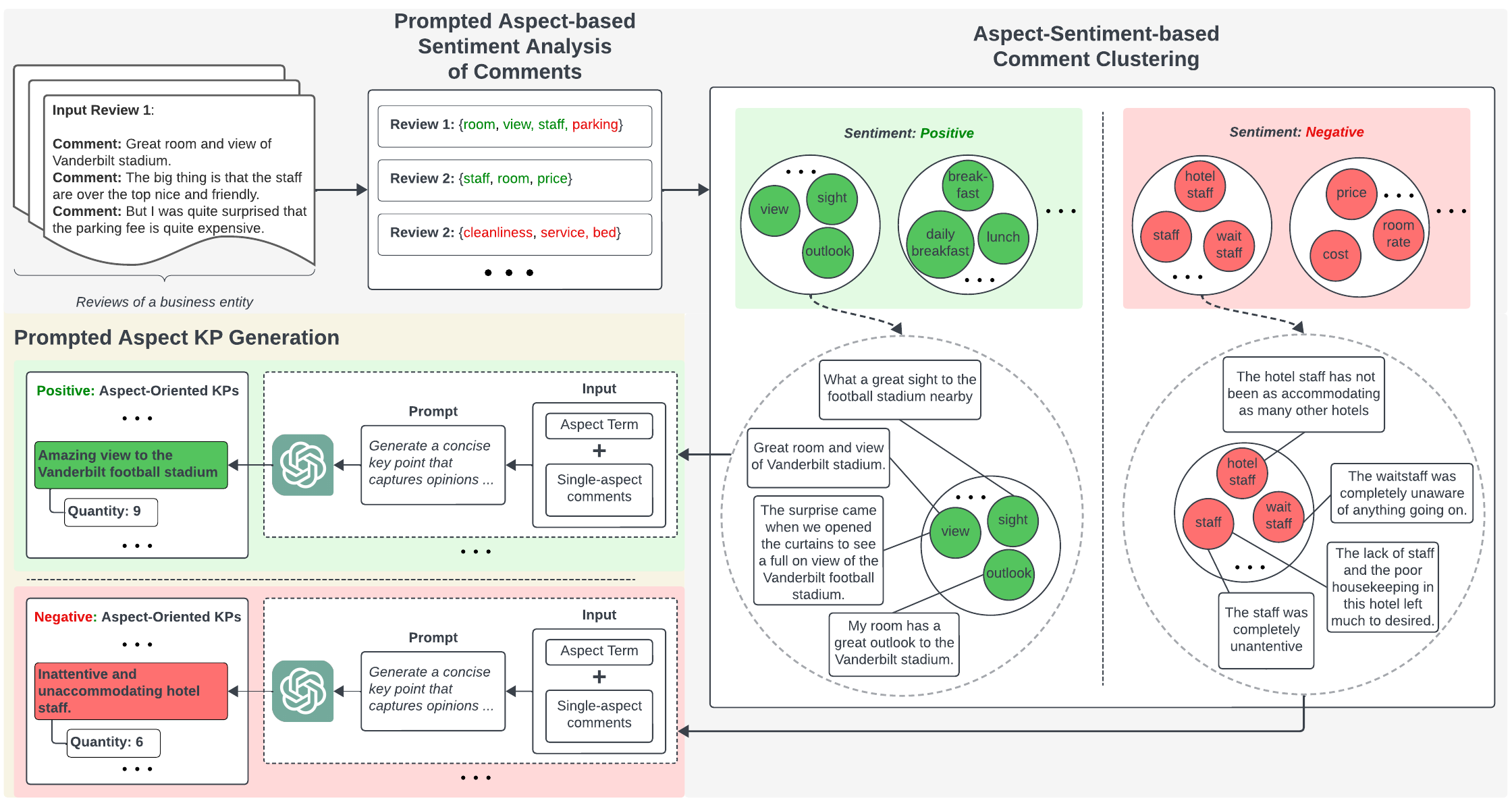}
    \caption{The PAKPA framework}
    \label{fig:PAKPA_Framework}
\end{figure*}

\section{Methodology}
\label{sec:methodology}

\begin{table*}[!ht]
    \centering
    \small
    \aboverulesep = 0pt
    \belowrulesep = 0pt
    \begin{tabularx}{1\textwidth}{|X|X|}
        \hline
        \textbf{Prompt for ABSA of Comments} & \textbf{Prompt for Aspect Key Point Generation} \\ \hline
        \makecell{
\begin{lstlisting}[frame=none,  aboveskip=0pt,belowskip=0pt, basicstyle=\tiny]
You will be provided with a review sentence delimited by triple quotes.
A review sentence usually covers the customer opinions expressed on different aspects of a product or service.

You are tasked to perform Aspect-based Sentiment Analysis to extract the user sentiments expressed on different aspects in the review.
Formally, we define subtask of extracting the aspects it corresponding sentiments as Aspect Extraction and Aspect Sentiment Classification:
- Aspect Extraction: Identifying aspect targets in opinionated text, i.e., in detecting the specific aspects of a product or service the opinion holder is either praising or complaining about. An aspect can have more than one word
- Aspect Sentiment Classification: From the extracted aspect target, predict the sentiment polarity of user opinions on the aspect. The sentiment polarity value can be: "positive", "neutral", and "negative".

Provide the answer in JSON format with the following keys: aspect, sentiment
\end{lstlisting}} & 
        \makecell{
\begin{lstlisting}[frame=none, aboveskip=0pt,belowskip=0pt, basicstyle=\tiny]
You will be provided with a list of user review comments delimited by triple quotes, and a list of common aspects shared by those reviews delimited by triple quotes
The comments in the list has been clustered by some common aspects and sentiment.
You are guided to generate a concise key point that captures opinions on the most popular aspect across the input comments, and also accomodate the provided list of common aspect.
Note that the generated key points must describe the opinion in only ONE aspect only and must not discuss multiple aspects. The generated key points must have 3-5 tokens.

Perform the following actions to solve this task:
- Identify the single and general aspect (e.g. atmosphere) that are common across the input aspects terms
- On the identified aspect, find the salient points of opinions mentioning that aspect across the input comments
Some invalid examples of key points with multiple aspects that must be avoided:
- "Enjoyable atmosphere with great music and live entertainment.", rather it should be "The atmosphere is very enjoyable."
- "Excellent wine selection and enjoyable atmosphere.", rather it should be "The wine selection is great."
\end{lstlisting}} \\
        \hline
    \end{tabularx}
    \caption{\label{table:prompt_table}
    Prompts for ``ABSA of Comments'' and ``Aspect Key Point Generation'' of the PAKPA framework. Full prompts with few-shot examples are provided in Appendix~\ref{sec:gpt3.5_prompt}}
\end{table*}

Figure~\ref{fig:PAKPA_Framework} illustrates our PAKPA framework with examples.
Given reviews for a business entity, 
PAKPA performs KPA for reviews and 
generates KPs of distinctive aspects and quantities measuring the prevalence of KPs. PAKPA consists of three components: 
\begin{itemize}
\item \emph{Prompted Aspect-Based Sentiment Analysis (ABSA) of comments:} extracts the aspect terms and sentiment -- positive or negative -- for each review comment (sentence),
\item \emph{Aspect sentiment-based comment clustering:} clusters comments sharing similar aspects and sentiments
\item \emph{Prompted aspect KP generation:} generates aspect KPs from comment clusters.
\end{itemize}
Core to our framework is to employ ABSA of review comments
to identify aspect terms in reviews and predict their sentiment,
which sets the basis for clustering comments based on aspects 
and for further generation of aspect-oriented KPs.
This idea is inspired by the early aspect-based structured summarization studies~~\citep{hu2004mining,ding2008holistic}, 
which aggregates review comments by their sentiment toward common aspects for more accurate quantification of opinions.
Importantly, prompted in-context learning strategies are employed for aspect-based sentiment analysis of review comments, and aspect-oriented KP generation and quantification.

\subsection{Prompted Aspect-based Sentiment Analysis of Comments}
\label{sec:prompted_abasa}
We designed and employed prompted in-context learning for LLMs to extract ABSA from reviews.
The task is to predict $(a,s)$ pairs -- \textbf{$(a)$}spect term, and \textbf{$(s)$}entiment (positive, neutral or negative) -- for each review sentence.
We developed a simple prompting strategy based on OpenAI~\footnote{\url{https://platform.openai.com/docs/guides/prompt-engineering}}'s prompt engineering guidelines.
Our prompts are structured into five parts, as shown in Table~\ref{table:prompt_table}: 1) Context of the review comment to be analyzed; 2) Definition of the ABSA task and the expected elements to retrieve; 3) Request for the LLM to provide the label in a JSON format; 4) Few-shot (18) examples to guide the LLM to generate the desired type of response; and 5) Review comment for ABSA predictions.
Experiments (Section~4.3) show that our prompted approach achieved reasonable performance on the aspect extraction and sentiment prediction tasks compared to supervised ABSA models.
\subsection{Aspect Sentiment-based Comment Clustering}
\label{sec:aspect_clustering}

Clustering comments directly based on their identical aspect terms can be highly overlapping 
because there are semantically similar aspect terms among the clusters.
We aim to construct clusters of comments such that comments of the same cluster will share the same aspect and sentiment, and each cluster has distinct aspect and sentiment from the other. 
To achieve this object, we leverage the (aspect, sentiment) pairs extracted from comments by our Prompted ABSA process (Section 3.1).
We propose a greedy algorithm to construct clusters for comments,
based on their sentiment and semantically similar aspect terms.

Let \mbox{$R_e = \{r_i\}_{i=1}^{|R_e|}$} denotes a set of review comments on a business entity $e$.
First we start by applying prompted ABSA (discussed in Section~\ref{sec:prompted_abasa}) on $r$ to extract possible ($a$)spect terms and 
the ($s$)entiment in a comment as a list of ($a$, $s$) pairs. Formally, this can be defined as \mbox{$O_r = \{(a_m, s_m)\}_{m=1}^{|O_r|}$}, where $s_m$ is the sentiment polarity of the $m$-th aspect in $r$. (\textsl{positive}, \textsl{neutral}, or \textsl{negative}).
Note that hereafter we filter all neutral sentiments in $O_r$.
We then aggregate all aspect terms ($a_m$) of the same sentiment in $r_i \in R_e$ into $A_{pol}$, where $pol$ is either the positive or negative.

Given a $A_{pol}$ of $R_e$, we first rank all aspects by descending order of their frequency in $R_e$.
Then we start with an empty $\mathbf{C}$, and iterate through every aspect in $A_{pol}$. 
For every aspect, we further iterate through every existing cluster in $\mathbf{C}$ and calculate the average 
cosine similarity score
to all included aspects of the cluster. 
Finally, only the aspect with the highest average cosine similarity score is added to the cluster with a threshold ($\lambda$) above 0.55, otherwise, a new cluster is created.
As shown in Figure~\ref{fig:PAKPA_Framework}, an example of semantically similar aspect terms is \emph{view}, \emph{sight}, and \emph{outlook}, 
which can be grouped into a cluster. 

We employ SpaCy~\citep{honnibal:2020}
to calculate the cosine similarity between aspect terms to form clusters.
Finally, comments sharing similar aspects, now grouped into clusters, are aggregated to become the input for the upcoming KP Generation stage, 
and the size of clusters is the quantity measuring the prevalence of KPs.

\subsection{Prompted Aspect-oriented KP Generation}
Unlike existing studies that rely on supervised 
text generation~\citep{li-etal-2023-hear},
we achieve Key Point Generation (KPG) by 
prompting an LLM to generate concise, distinct KPs 
from clusters of comments with the semantically similar aspect terms.
Our main idea is that semantically similar aspect terms of a cluster of comments can be a good signal to infer a high-level and more general aspect-oriented textual description as the KP. 
Specifically, we designed the prompt for Aspect KPG based on simple prompting strategies suggested by the OpenAI prompt engineering guideline
~\footnote{\url{https://platform.openai.com/docs/guides/prompt-engineering}} to write clear instructions to prompt the model.
Our prompt is structured into six parts, as shown in Table~\ref{table:prompt_table}: 1) Context of the KPG input to be summarized; 2) Definition of the Aspect KPG task and the output requirement; 3) Summarization steps to guide the LLM to infer the general aspects from the cluster's aspect terms and then generate aspect-oriented KP; 4) One-shot example to guide the LLM to generate the desired type of response; 5) Guiding the LLM through invalid generation examples to avoid, along with preferred correction for practicing; and 6) KPG input for summarization.
We provide details of the prompt on LLMs for aspect-based KPG in Listing~\ref{lst:gpt_kpg_1_shot_CoT} (Appendix~\ref{sec:gpt3.5_prompt}).

\section{Experiments}

\subsection{Implementation Details and Baselines}
\label{sec:baseline}

PAKPA was implemented with different LLMs, including open-source models \texttt{Vicuna-7B}, \texttt{Mistral-7B}, and the commercial model \texttt{GPT3.5}.
Note that we employed open-source LLMs throughout PAKPA for both the ABSA stage and the KP generation stage, but we did not apply \texttt{GPT3.5} for the ABSA stage due to the exorbitant cost for thousands of reviews as the input context.
We benchmark PAKPA against various baselines for extractive KPA, abstractive KPA, and the recent prompted opinion summarization. 
\paragraph{Extractive KPA:} We compare PAKPA against two latest extractive KPA systems \textbf{RKPA-Base}~\citep{bar-haim-etal-2021-every} and \textbf{ABKPA}~\citep{2024aspectbased}.
RKPA-Base
is the first extractive KPA system for review summarization.
It leverages a quality ranking model~\citet{gretz2020large} to select KP candidates, and integrates sentiment analysis and collective key point mining into matching 
comments to the extracted KPs.
ABKPA integrates
 ABSA into extracting and matching of KPs to comments 
for more precise matching and quantification of key points.
We implement all models based on their default settings.

\paragraph{Abstractive KPA:} We implemented 
two latest abstractive KPA systems \textbf{Enigma+}~\citep{kapadnis-etal-2021-team} and $\textbf{SKPM}_{\textbf{Base}}\textbf{(IC)+}$~\citep{li-etal-2023-hear}.
Enigma+ is adapted from the original Enigma framework to review data, which uses a Pegasus~\citep{zhang2020pegasus} summarization model to 
generate KPs from comments,
and selects the top 40 summaries based on their ROUGE scores.
Similarly, ${\rm SKPM_{\rm Base}}(\rm IC)+$ is adapted for reviews,~\footnote{We reproduced this model based on the best configuration provided.} employing \texttt{BERTopic}~\citep{grootendorst2022bertopic} to cluster sentences and \texttt{Flan-T5}~\citep{chung2022scaling} to generate KPs.
To fully adapt these works from arguments to reviews, we replace the topic and stance attribute in the input with business category and sentiment.
We fine-tune all models using an annotated KP Matching dataset~\citep{2024aspectbased}.

All above baselines were implemented either using the PyTorch module or the Huggingface transformers framework, and were trained on an NVIDIA GeForce RTX 3080Ti GPU.

\paragraph{Prompted Opinion Summarization:} 
To evaluate the utility of KPA systems for textual summaries, 
we also compare them against 
the latest prompted opinion summarization model \textbf{Recursive GPT3-Chunking (CG)}~\citep{bhaskar2023prompted}, which 
recursively chunks and prompts 
\texttt{GPT3.5} to generate textual summaries from user reviews.
The final summary from this baseline is 
a paragraph rather than a list of KPs. 
For fair comparison, we follow the strategy of~\citet{bhaskar2023prompted} by again prompting \texttt{GPT3.5} 
to split and rephrase the summary sentences into KPs.~\footnote{Also known as the atomic value judgement~\citep{bhaskar2023prompted}.}

\subsection{Datasets and Evaluation Dimensions}
\paragraph{Datasets}
To evaluate both the textual quality and prevalence precision for KPs,
we consider two popular datasets on business reviews, namely \textsc{Space} and \textsc{Yelp}. 
\textbf{(1) \textsc{Space}}, featuring TripAdvisor hotel reviews, stands out as the only dataset providing human-annotated aspect-specific summaries and, therefore, serves as an ideal ground truth for evaluating our aspect-based generation of KPs in PAKPA.
The dataset facilitates the evaluation of KP quality in capturing the main viewpoints of users across various aspects (e.g., location and cleanliness).
\textbf{(2) \textsc{Yelp}} is a widely used dataset for review summarization including a wider variety of business categories.
This dataset is used to evaluate both the textual 
quality and quantification performance of KPs.
Details of the datasets can be found in Appendix~\ref{sec:dataset_details}.

\paragraph{Evaluation of KP Textual Quality with Aspect-Specific Ground Truth}
\label{sec:automatic_evaluation}
\textsc{Space} provides the reference summaries for this evaluation.
Positive and negative summaries are evaluated separately.~\footnote{we use SpaCy to perform sentiment analysis on every referenced summary sentence.}
We first perform a lexical comparison between generated KPs and the ground truth 
by computing the highest ROUGE score between generated and reference key points for each business entity and then average the maxima.
Neverthless, KPs generated from abstractive KPA systems should not only be evaluated based on lexical similarity against ground truth summaries. 
We, therefore, employ the ~\emph{set-level KPG evaluation}~\citep{li-etal-2023-hear}, which explicitly measures the quality 
between two sets of generated and reference KPs based on their semantic similarity.
For all business entities, 
we calculate the semantic similarity scores between the corresponding group of prediction and reference before macro-averaging their values 
to obtain \emph{Soft-Precision (sP)} and \emph{Soft-Recall (sR)}.
While $sP$ finds the reference KP with the highest similarity score for each generated KP, $sR$ is vice-versa.
We further define \emph{Soft-F1} ($sF1$) as the harmonic mean between $sP$ and $sR$ as below, 
where $f$ computes similarities between two individual key points, $\mathcal A$, $\mathcal{B}$ is the set of candidates and references and $n=|\mathcal{A}|$ and $m=|\mathcal{B}|$, respectively. 
\begin{equation}
    \small
    sP = \frac{1}{n} \times \sum_{ \alpha_i\in\mathcal{A}} \max_{\beta_j\in\mathcal{B}} f(\alpha_{i},  \beta_{j})
\end{equation}
\begin{equation}
    \small
    sR = \frac{1}{m} \times \sum_{ \beta_i\in\mathcal{B}} \max_{\alpha_j\in\mathcal{A}} f(\alpha_{i},  \beta_{j})
\end{equation}

We use state-of-the-art semantic similarity evaluation methods  BLEURT~\citep{sellam-etal-2020-bleurt} and BARTScore~\citep{yuan2021bartscore} as $f_{max}$.
For fair comparison, we select only KPs of at least 15 matched comments~\footnote{approximately equivalent to the top 7-10 KPs with the highest prevalence across the models for each business.}.

\paragraph{Evaluation of KP Faithfulness and Information Quality}
We manually evaluated the information quality of generated KPs considering 7 different dimensions, divided into two groups.
The first group, inspired by previous KPA works~\citep{friedman-etal-2021-overview,li-etal-2023-hear}, evaluates how well the generated KPs summarize the salient information from the corpus.
It assesses KPs
based on criteria \textsc{Redundancy}, \textsc{Coverage}, and \textsc{Faithfulness} (contrary to hallucination). 
The second group measures the utility of 
generated KPs for summarization,
under four dimensions~\citep{bar-haim-etal-2021-every}:
\textsc{Validity}, \textsc{Sentiment}, \textsc{Informativeness} and \textsc{Single\ Aspect}.
Details of these dimensions are in Appendix~\ref{sec:kp_quality_dimensions}.

We conducted pairwise comparison of KPs from different systems using Amazon Mechanical Turk (MTurk). Given a dimension for evaluation, each comparison involved choosing the better one from two sets of KPs, each taken from a different system. We selected the top 5 KPs by prevalence for each sentiment. Using the Bradley-Terry model~\citet{friedman-etal-2021-overview}, we calculated rankings from these comparisons among the models. We ensured high-quality annotations by employing workers with an approval rate of 80\% or higher and at least 10 approved tasks, while hiding ABSA details and framework identities to prevent bias. For an example of an annotation, see Appendix~\ref{sec:pairwise_kp_quality_annotation_guideline}.
We only performed this evaluation on the \textsc{Yelp} dataset, as it contains reviews for five business categories, including hotel reviews of \textsc{Space}. 
Note also that 
to maintain a reasonable annotation cost, for every category in \textsc{Yelp}, we select only one top popular business entity with the highest average number of KPs being generated across the models.

\paragraph{Evaluation of KP Quantification Precision}
In this experiment, we evaluate the precision of different systems for matching KPs to comments to measure the prevalence of KPs, namely   %
the KP quantification precision~\citep{bar-haim-etal-2021-every}.
Following previous studies~\citep{bar-haim-etal-2021-every,2024aspectbased}, 
this was conducted on \textsc{Yelp} for various business categories. 
Adjustments were made to some KPA baselines (e.g., RKPA-Base, ABKPA, Engima+) to ensure comparable \emph{review coverage}~\citep{bar-haim-etal-2021-every}~\footnote{Fraction of comments captured and quantified in the summary} by setting an appropriate threshold ($t_{match}$) for selecting the best-matching comment-KP pairs.
For annotation, we employed 6 MTurk crowd workers per comment-KP pair, selecting only those with an 80\% or higher approval rate and at least 10 approved tasks. 
Following Bar-Haim et al.'s, we exclude annotators with Annotator-$\kappa <0$ for quality control.
This score averages all pairwise Cohen's Kappa~\citep{landis1977measurement} for a given annotator, for any annotator sharing at least $50$ judgments with at least $5$ other annotators.
For labelling correct matches, at least 60\% of the annotators had to agree that the match is correct, otherwise, it is incorrect.

\begin{table}[tbh]
    \centering
    \small
    \aboverulesep = 0pt
    \belowrulesep = 0pt
    \begin{tabularx}{0.5\textwidth}{|l|X|X|}
        \hline
        \textbf{Task} & \textbf{AE} & \textbf{ASC} \\ \hline
        \emph{Prompted \texttt{Vicuna7B}} & 80.5 & 77.14 \\
        \emph{Prompted \texttt{Mistral7B}} & 78.56 & 76.88 \\\hdashline
        \emph{Snippext (Full training)} & 79.65 & 80.45 \\
        \emph{Snippext (Low-resource)} & 77.18 & 77.4 \\\hline
    \end{tabularx}
\caption{\label{table:prompted_absa_benchmark_semeval}
    The F1 score of prompted LLMs and the SOTA Snippext model~\citep{miao2020snippext} for ABSA, evaluated on the Aspect Extraction (AE) and Aspect Sentiment Classification (ASC) tasks.
    }
\end{table}

\subsection{Results}

\paragraph{Evaluation of ABSA}
To evaluate the effectiveness of prompted LLMs compared to supervised approaches for ABSA, we benchmark their performance on two tasks, namely Aspect Extraction (AE) and Aspect Sentiment Classification (ASC), from SemEval 2016 Task 5~\citep{pontiki-etal-2016-semeval} and SemEval 2014 Task 4~\citep{pontiki-etal-2014-semeval} respectively.
Experimental results from Table~\ref{table:prompted_absa_benchmark_semeval}
show that prompted LLMs achieved reasonable performance on AE and ASC tasks compared to the state-of-the-art (SOTA) ABSA model Snippext~\citep{miao2020snippext}.

\begin{table*}[t]
    \centering
    \scalebox{0.9}{
    \begin{tabular}{lccccccccc}
    \toprule
    {} & \multicolumn{3}{c}{ROUGE} & \multicolumn{3}{c}{BARTScore} & \multicolumn{3}{c}{BLEURT}\\
    \cmidrule(r){2-4} \cmidrule(r){5-7} \cmidrule(r){8-10}
     & R-1 & R-2 & R-L & sP & sR & sF1 & sP & sR & sF1\\
    \midrule
    \emph{${\rm PAKPA_{Vicuna7B + GPT3.5}}$}  & \textbf{0.648} & \textbf{0.364} & \textbf{0.510} & 0.74 & 0.66 & 0.70 & \textbf{0.61} & \textbf{0.51} & \textbf{0.56} \\
    \emph{${\rm PAKPA_{Mistral7B + GPT3.5}}$}  & 0.588 & 0.341 & 0.453 & 0.73 & 0.64 & 0.68 & 0.58 & 0.50 & 0.54 \\
    \emph{${\rm PAKPA_{Mistral7B + Mistral7B}}$}  & 0.531 & 0.269 & 0.440 & 0.74 & \textbf{0.68} & \textbf{0.71} & 0.58 & 0.50 & 0.54 \\
    \emph{${\rm PAKPA_{Vicuna7B + Vicuna7B}}$}  & 0.515 & 0.231 & 0.371 & 0.73 & 0.66 & 0.70 & 0.61 & 0.49 & 0.54 \\
    \hdashline
    \emph{Enigma+}~\citep{kapadnis-etal-2021-team} & 0.628 & 0.346 & 0.492 & 0.74 & 0.65 & 0.69 & 0.56 & 0.49 & 0.52 \\
    \emph{CG}~\citep{bhaskar2023prompted} & 0.416 & 0.205 & 0.406 & 0.73 & 0.56 & 0.63 & 0.52 & 0.45 & 0.48 \\
    \emph{${\rm SKPM_{\rm Base}}(\rm IC)+$}~\citep{li-etal-2023-hear} & 0.335 & 0.139 & 0.318 & 0.67 & 0.58 & 0.62 & 0.38 & 0.36 & 0.37 \\
    \hdashline
    \emph{RKPA-Base}~\citep{bar-haim-etal-2021-every} & 0.552 & 0.292 & 0.488 & \textbf{0.75} & 0.59 & 0.66 & 0.59 & 0.46 & 0.52 \\
    \emph{ABKPA}~\citep{2024aspectbased} & 0.442 & 0.245 & 0.422 & 0.74 & 0.63 & 0.68 & 0.56 & 0.46 & 0.51 \\
    \bottomrule
    \end{tabular}
    }
    \caption{\label{table:automatic_evaluation}
    \textbf{
        (\textsc{Space}}) Textual quality evaluation of generated KPs with aspect-specific ground truth.
    sP, sR and sF1 refer to Soft-Precision, Soft-Recall, and Soft-F1 respectively based on set-level evaluation method.
    Statistical analysis on all metrics shows that PAKPA significantly outperforms the baselines (paired t-test $p << 0.05$).
    }
\end{table*}

\paragraph{Evaluation of KP Quality}

Table~\ref{table:automatic_evaluation} presents our evaluation of the textual quality of KPs generated by different systems, focusing on their lexical and semantic similarity to the SPACE ground truth. 
In its best setting, our framework, PAKPA, outperforms other baselines across all metrics, capturing approximately 66\% (sR = 0.66) of the viewpoints expressed in manually annotated aspect-specific summaries. 
Notably, SKPMBase(IC)+, despite its superiority over Enigma+ in argument summarization~\citep{li-etal-2023-hear}, underperforms in generating quality KPs from reviews, as indicated by most metrics. This inferiority is attributed to SKPMBase(IC)+'s vulnerability to hallucination when summarizing from a large set of comments, due to its reliance on limited supervised training data. 
Conversely, Enigma+, which generates KPs by rephrasing a single review sentence, maintains acceptable quality in its abstractive KP generation.

\begin{table*}[htbp]
    \centering
    \scalebox{0.9}
    {
    \begin{tabular}{lccc:cccc}
        \toprule
        & CV & FF & RD & VL & SN & IN & SA\\
        \midrule
        \emph{${\rm PAKPA_{Vicuna7B + GPT3.5}}$} & \textbf{28.44} & \textbf{26.56} & \textbf{25.34} & \textbf{35.23} & \textbf{31.11} & \textbf{25.9} & \textbf{24.8} \\
        \emph{Enigma+}~\citep{kapadnis-etal-2021-team} & 11.06 & 11.17 & 14.7 & 9.99 & 9.54 & 13.49 & 17.52 \\
        \emph{CG}~\citep{bhaskar2023prompted} & 15.12 & 12.84 & 15.73 & 10.36 & 14.6 & 12.59 & 10.79 \\
        ${\rm SKPM_{\rm Base}}(\rm IC)+$~\citep{li-etal-2023-hear} & 9.94 & 12.41 & 13.28 & 7.7 & 8.87 & 13.04 & 9.34\\ \hdashline
        \emph{RKPA-Base}~\citep{bar-haim-etal-2021-every} & 16.20 & 22.28 & 15.73 & 22.91 & 20.75 & 21.02 & 18.77 \\
        \emph{ABKPA}~\citep{2024aspectbased} & 19.24 & 14.74 & 15.21 & 13.81 & 15.12 & 13.96 & 18.77\\
        \bottomrule
    \end{tabular}
    }
    \caption{(\textbf{\textsc{Yelp}}) Information quality evaluation of generated KPs by different dimensions.
    Reported are the Bradley Terry scores of 7 dimensions, from left to right, \textsc{Coverage}, \textsc{Faithfulness} and \textsc{Redundancy}, \textsc{Validity}, \textsc{Sentiment}, \textsc{Informativeness}, \textsc{SingleAspect}. A visual overview can also be found in Figure~\ref{fig:kp_quality_eval_results_fig} (Appendix~\ref{sec:kp_quality_visual_overview})
    \label{table:kp_quality_eval_results_T}}
\end{table*}
Our manual evaluation~\footnote{To maintain reasonable annotation cost, we only conducted manual evaluation on the best LLM configuration for PAKPA (${\rm PAKPA_{Vicuna7B + GPT3.5}}$), selected from Table~\ref{table:automatic_evaluation}.}
on KP information quality further supports above findings.
Table~\ref{table:kp_quality_eval_results_T} 
highlights the Bradley Terry scores, measured by 7 information quality dimensions, of the KPs produced on \textsc{Yelp}.
Overall, 
on all 7 dimensions,
PAKPA exhibits the highest and most stable performance.
For summarizing the salient points, 
our framework outperforms other baselines significantly on \textsc{Coverage} (CV) and \textsc{Redundancy} (RD), 
as it suggests that our approach captures more diverse opinions 
and also more effectively reduces redundancy in the KPs thanks to its aspect-based clustering and generation process.
Importantly, PAKPA outperforms all baselines in \textsc{Faithfulness}, more than doubling the effectiveness in reducing hallucinations compared to other abstractive summarization systems.
For generating good KPs for reviews, PAKPA outperforms other baselines greatly on \textsc{Validity} (VL), 
mainly because our approach uses LLMs to generate KPs that are aligned better with the expected format.
Nevertheless, high scores SN, IN and SA also also shows that PAKPA can generate KPs with richful opinion information, expressing clearer sentiment and on more specific aspect than other baselines.
\begin{table*}[!ht]
    \centering
    \scalebox{0.9}
    {
    \begin{tabular}{lccccc:c}
        \toprule
         & \textbf{Arts} & \textbf{Auto} & \textbf{Beauty} & \textbf{Hotels} & \textbf{Rest} & \textbf{Avg.}\\
        \midrule
        \emph{${\rm PAKPA_{Vicuna7B + GPT3.5}}$} & \textbf{0.98} & \textbf{0.93} & \textbf{0.96} & \textbf{0.94} & 
        \textbf{0.94} & \textbf{0.95} \\
        \emph{ABKPA} & 0.80 & 0.86 & 0.80 & 0.86 & 0.82 & 0.83 \\
        ${\rm SKPM_{\rm Base}}(\rm IC)+$ & 0.80 & 0.79 & 0.73 & 0.77 & 0.70 & 0.76 \\
        \emph{RKPA-Base} & 0.62 & 0.63 & 0.63 & 0.69 & 0.71 & 0.66 \\
        \emph{Enigma+} & 0.61 & 0.69 & 0.58 & 0.55 & 0.69 & 0.64 \\
        \bottomrule
    \end{tabular}
    }
    \caption{\label{table:quantification_precision}
    (\textbf{\textsc{Yelp}}) Quantification precision evaluation of generated KPs.
    The precision is reported on five business categories: \textbf{Arts} (\& Entertainment), \textbf{Auto}(motive), \textbf{Beauty} (\& Spas), \textbf{Hotels}, \textbf{Rest}(aurants).
    }
\end{table*}
\paragraph{Performance of PAKPA}
Table~\ref{table:automatic_evaluation} additionally 
presents the performance of PAKPA using different base LLMs on the \textsc{Space} dataset.
Overall, implementing a combination of LLMs as base models for PAKPA leads to signficantly better performance than using an open-source LLM alone.
Specifically, among multi-LLMs configurations, (\texttt{Vicuna7B + GPT3.5}) achieves state-of-the-art performance, mainly due to the powerful generative capability of \texttt{GPT3.5} on KP Generation.
On the other hand, for configurations based on one LLM, (\texttt{Mistral7B + Mistral7B}) outperforms (\texttt{Vicuna7B + Vicuna7B}). 
It is important to note that although \texttt{Mistral7B} outperforms \texttt{Vicuna7B} on KP Generation task, \texttt{Vicuna7B} is still the top performer of the ABSA task, as shown in Table~\ref{table:prompted_absa_benchmark_semeval}.
This crucially contributes to the state-of-the-art performance of (\texttt{Vicuna7B + GPT3.5}).
\paragraph{Evaluation of KP Quantification Precision using \textsc{Yelp}}
Table~\ref{table:quantification_precision} presents the precision scores for all KPA models, which 
shows their general performance of 
matching input comments to the generated KPs across 5 business categories of \textsc{Yelp}.
Overall, PAKPA outperforms all baselines, with improvements of up to $31\%$ in the matching precision score, and the performance is stable across the business categories.
RKPA-Base, Enigma+ and ${\rm SKPM_{\rm Base}}(\rm IC)+$, 
without access 
to the ABSA information of reviews to create aspect-specific summaries, show an inferior quantification performance compared to ABKPA and PAKPA.
Integrating ABSA into the KPA system, either in extractive or abstractive techniques,  then becomes a critical factor for achieving state-of-the-art performance for review summarization.
For example, ${\rm SKPM_{\rm Base}}(\rm IC)+$, whose architecture was proven effective on argument debates, achieves inferior performance when applied for reviews compared with ABKPA, an extractive KPA system incorporating ABSA.
It is also worth noting that 
previous KPA studies with abstractive implementation, 
though committed to generating more concise yet less redundant KPs,
always have inferior matching performance to the SOTA extractive techniques.
More specifically, in most business categories, Enigma+, an early KPA system applying abstractive summarization, is outpaced by RKPA-Base, an early extractive system.
Such inferiority is primarily due to 
data scarcity issues 
for fine-tuning pre-trained language models (PLM) to generate high-quality KPs for reviews, making existing abstractive KPA frameworks prone to hallucination.
Interestingly, our abstractive aspect-based PAKPA system outperforms the extractive aspect-based system ABKPA, 
largely due to the utility of our prompted in-context learning on LLMs and aspect-oriented KP generation approach.

\paragraph{Error Analysis}
By analyzing the errors in KP generation of our system 
across business categories and datasets,
we found several systematic patterns of errors.
A frequent error type is KPs containing extra information related to its main aspects.
An example KP in this category is 
``Overpriced breakfast with mediocre coffee''.
This sometimes happens when more specific aspect terms (e.g., ``coffee'') are clustered with more general ones (e.g. ``breakfast''), and they cover different opinion information that is difficult to generalize. 
In some other cases, 
KPs generated for a cluster can also be overly generalized, and so  coverage includes 
the major opinions of comments but may ignore the minor ones.
For example, the comment ``I love their pastries and they have a decent selection of yummy cookies.'' was matched to the aspect ``Delicious and diverse cake options'',  
which should also be referred to as the ``bread'' aspect.

\subsection{Case Studies}
\begin{table}[!ht]
    \centering
    \small
    \aboverulesep = 0pt
    \belowrulesep = 0pt
    \begin{tabularx}{0.5\textwidth}{|m{4.3em}|X|}
        \toprule
        {} & \textbf{Key Points}\\ \hline
        \emph{PAKPA} & Poor service and unresponsive staff. \\ \hline
        ${\rm SKPM_{\rm Base}}$-$(\rm IC)+$ & didn't work at all - the front desk staff was rude, rude, and!!! \\ \hline
        \emph{Enigma+} & They don't listen!!!! \\ \hline
        \emph{ABKPA} & Overall unprofessional and unorganized. \\ \hline
        \emph{RKPA-Base} & are rude, slow and disrespectful. \\ \hline
        \emph{CG} & However, negative aspects mentioned included \textbf{issues with room conditions}, \textbf{slow service}, \textbf{noise}, \textbf{safety concerns}, and \textbf{lack of amenities}. \\
        \bottomrule
    \end{tabularx}
    \caption{\label{table:one_kp_example}
    KPs generated by different KPA systems summarizing a ``Hotel'' business of \textsc{Yelp}}
\end{table}

We conduct case studies to evaluate the redundancy and hallucination of generated KPs for a ``Hotel'' business of \textsc{Yelp}, as shown in Table~\ref{table:one_kp_example}.
Overall, PAKPA stands out for generating KPs with minimal redundancy, 
also being highly informative and at good aspect diversity (e.g., ``Poor service and unresponsive staff.''),
which is superior to previous abstractive counterparts such as ${\rm SKPM_{\rm Base}}(\rm IC)+$ or Enigma+ that tend to produce 
repetitive, hallucinated and overly broad KPs (e.g., ``didn't work at all - the front desk staff was rude, rude, and!!'', ``They don't listen!!!!'').
Furthermore, the RKPA-Base and ABKPA models
still cannot provide KPs covering sufficient aspect information and as valid and fluent as PAKPA (e.g., ``Overall unprofessional and unorganized.'', ``are rude, slow and disrespectful.'').
More generated KP samples can be found in Table~\ref{table:top_negative_kps} and~\ref{table:top_positive_kps} (Appendix~\ref{sec:kp_summary_examples}).

\section{Conclusion}
In this paper, we propose Prompted Aspect Key Point Analysis (PAKPA), a novel KPA framework applying abstractive summarization for opinion quantification.
PAKPA addresses the 
issues of KPs with overlapping opinions, hallucination, and inaccurate quantification
of previous sentence-based 
KPA approaches.
Compared with previous studies, our approach effectively makes use of ABSA in business reviews 
to generate KPs grounded in aspects and achieve more accurate quantification. 
Experimental results show that our solution greatly enhances both the quantitative performance and quality of KPs.
Secondly, our prompted in-context learning approach  
also deviates from the conventional supervised learning approach and removes the need for large amounts of annotated data for supervised training and fine-tuning. 
\section*{Acknowledgement}
This research is supported in part by the Australian Research Council Discovery Project DP200101441.

\section*{Limitations}

We evaluated the textual quality of aspect KPs only on SPACE, as it is the only (to our best knowledge) public dataset with ground-truth human-annotated aspect-oriented textual summaries.

\section*{Ethics Statement}
We have applied ethical research standards in our organization for data collection and processing throughout our work.

The \textsc{Yelp} dataset used in our experiments was officially released by Yelp, while the \textsc{Space} dataset was publicly crowdsourced and released by the research publication for benchmarking opinion summarization framework.
Both datasets was published by following their ethical standard, after removing all personal information.
The summaries do not contain contents that are harmful to readers.

We ensured fair compensation for crowd annotators on Amazon Mechanical Turk. We setup and conducted fair payment to workers on their annotation tasks/assignments according to our organization's standards, with an estimation of the difficulty and expected time required per task based on our own experience. Especially, we also made bonus rewards to annotators who exerted high-quality annotations in their assignments.

\bibliography{anthology,custom}
\bibliographystyle{acl_natbib}

\appendix

\section{Prompts for \texttt{GPT3.5}}
\label{sec:gpt3.5_prompt}
We present the zero-shot and few-shot prompts for Aspect-Based Sentiment Analysis (ABSA) and Aspect-based Key Point Generation in Listing~\ref{lst:gpt_absa_18_shot} and~\ref{lst:gpt_kpg_1_shot_CoT}.

\begin{figure*}
    \centering
\begin{lstlisting}[caption={Few-shot prompt (18 examples) for prompting \texttt{GPT3.5} on fine-grained Aspect-based sentiment analysis. Please refer to our released code for full prompts.}, basicstyle=\small, label=lst:gpt_absa_18_shot]
You will be provided with a review sentence delimited by triple quotes.
A review sentence usually covers the customer opinions expressed on different aspects of a product or service.

You were tasked to perform Aspect-based Sentiment Analysis to extract the user sentiments expressed on different aspects in the review.
Formally, we define subtask of extracting the aspects it corresponding sentiments as Aspect Extraction and Aspect Sentiment Classification:
- Aspect Extraction: Identifying aspect targets in opinionated text, i.e., in detecting the specific aspects of a product or service the opinion holder is either praising or complaining about. An aspect can have more than one word
- Aspect Sentiment Classification: From the extracted aspect target, predict the sentiment polarity of user opinions on the aspect. The sentiment polarity value can be: "positive", "neutral", and "negative".

Provide the answer in JSON format with the following keys: aspect, sentiment

Review sentence: \"\"\"Movies cost $ 14 , and there is no student discount at this location .\"\"\"
Answer: [{'aspect': 'student discount', 'sentiment': 'negative'}]

Review sentence: \"\"\"Our tour guide was knowledgeable about the property and about all things Frank Lloyd Wright .\"\"\"
Answer: [{'aspect': 'tour guide', 'sentiment': 'positive'}]

Review sentence: \"\"\"BMW Henderson made my purchase easy and stress free .\"\"\"
Answer: [{'aspect': 'purchase', 'sentiment': 'positive'}]

Review sentence: \"\"\"I had a male therapist and he was amazing !\"\"\"
Answer: [{'aspect': 'male therapist', 'sentiment': 'positive'}]

...

Review sentence: \"\"\"Be sure to accompany your food with one of their fresh juice concoctions .\"\"\"
Answer: [{'aspect': 'food', 'sentiment': 'neutral'}, {'aspect': 'fresh juice concoctions', 'sentiment': 'positive'}]

Review sentence: \"\"\"During busy hrs, i recommend that you make a reservation .\"\"\"
Answer: [{'aspect': 'reservation', 'sentiment': 'neutral'}]

Review sentence: \"\"\"The menu, which changes seasonally, shows both regional and international influences .\"\"\"
Answer: [{'aspect': 'menu', 'sentiment': 'neutral'}]

Review sentence: \"\"\"Our waitress had apparently never tried any of the food, and there was no one to recommend any wine.\"\"\"
Answer: [{'aspect': 'waitress', 'sentiment': 'negative'}, {'aspect': 'food', 'sentiment': 'neutral'}, {'aspect': 'wine', 'sentiment': 'neutral'}]
"""
\end{lstlisting}
\end{figure*}

\begin{figure*}
    \centering
\begin{lstlisting}[caption={One-shot prompt for prompting \texttt{GPT3.5} on KP Generation.}, basicstyle=\small, label=lst:gpt_kpg_1_shot_CoT]
You will be provided with a list of user review comments delimited by triple quotes, and a list of common aspects shared by those reviews delimited by triple quotes
The comments in the list has been clustered by some common aspects and sentiment.
You were guided to generate a concise key point that captures opinions on the most popular aspect across the input comments, and also accomodate the provided list of common aspect.
Note that the generated key points must describe the opinion in only ONE aspect only and must not discuss multiple aspects. The generated key points must have 3-5 tokens.

Perform the following actions to solve this task:
- Identify the single and general aspect (e.g. atmosphere) that are common across the input aspects terms
- On the identified aspect, find the salient points of opinions mentioning that aspect across the input comments
Some invalid examples of key points with multiple aspects that must be avoided:
- "Enjoyable atmosphere with great music and live entertainment.", rather it should be "The atmosphere is very enjoyable."
- "Excellent wine selection and enjoyable atmosphere.", rather it should be "The wine selection is great."

Comments: """['The bartenders were so sweet and were very responsive .', 'The staff is fantastic and responsive .', 'The staff was so accommodating and kind !', 'The hotel staff went above and beyond with their customer service .', 'The staff was super accommodating and made planning a cinch .', 'Front desk staff was welcoming and accommodating .', 'All staff were friendly , helpful & professional .  ', 'Everyone of the staff has been super friendly and accommodating .', 'Rooms are comfortable and staff are friendly .', 'The staff was courteous & informative .', 'Mandatory valet parking with excellently quick service and attentive desk staff .', 'Much better location and competent staff !', 'The staff is amazing - upbeat , involved , and made great recommendations .  ', 'The front desk staff was unbelievably friendly and accommodating .', 'Clean , comfortable and friendly , accommodating staff .', 'Their service was professional , accommodating , fast and cordial .', 'The staff was friendly and rectified any mistakes on our reservation .', 'The front staff is accommodating , informative , and friendly .   ', 'The staff was courteous and efficient .', 'The staff was friendly and courteous .', 'Pool , spa , gym -- super courteous staff , what more could you want ?']"""
Aspects: """['bartenders', 'staff', 'hotel staff', 'front desk staff', 'desk staff', 'front staff']"""
Key Point: Friendly and helpful staff .
\end{lstlisting}
\end{figure*}

\section{Details of the Experimental Datasets}
\label{sec:dataset_details}
\paragraph{\textsc{Space}} 
A large-scale opinion summarization dataset built on TripAdvisor hotel reviews, with its test set containing a large collection of human-written summaries (for reviews of 50 hotels) usable as the ground truth in our experiment.
To our best knowledge, 
\textsc{Space} stands out as the sole dataset providing human-written aspect-specific summaries, serving as an ideal ground truth for evaluating our aspect-based generation of KPs in PAKPA.
In this experiment, we opt to select both the \emph{general summaries}, i.e., short and high-level overview of popular opinions, and \emph{aspect-specific summaries}, detail on individual aspects (e.g., location, cleanliness) of \textsc{Space} because they both can be represented by our KPs.
Note that we ignore the aspect label of these summaries and focus only on their content in our experiment.
To maintain a reasonable run time, we also limit the selection to only the top 10 hotels with the highest number of reviews in \textsc{Space}, and we exclude reviews with more than 15 sentences.
We show additional statistics of our \textsc{Space} dataset in Table~\ref{table:space_statistics}

\begin{table}
    \centering
    \small
    \caption{Statistics of \textsc{Space}}
    \label{table:space_statistics}
    \begin{tabularx}{0.5\textwidth}{|
        l|
        >{\centering\arraybackslash}X|
        >{\centering\arraybackslash}X|
        >{\centering\arraybackslash}X|
        >{\centering\arraybackslash}X|}
        \hline
            \textbf{Category} & \textbf{\# Reviews} & \textbf{\# Sentences} & \textbf{\# Sentences Per Review} & \# \textbf{Sentences Per Reference Summary} \\ \hline
            Hotels & 946 & 7510 & 7.94 & 2.48 \\
        \hline
    \end{tabularx}
\end{table}

\paragraph{\textsc{Yelp}} Business reviews from the Yelp Open Dataset~\footnote{\url{https://www.yelp.com/dataset}}, as being utilized in previous extractive KPA study for reviews~\citep{bar-haim-etal-2021-every,2024aspectbased}, 
targetting five business categories;
\emph{Arts \& Entertainment} (25k reviews), \emph{Automotive} (41k reviews), \emph{Beauty \& Spas} (72k reviews), \emph{Hotels} (8.6K reviews), and \emph{Restaurants} (680k reviews).
We applied additional filters and selections to the dataset to maintain a reasonable runtime as follows.
First, we excluded reviews with more than 15 sentences.
Second, on the remaining data, we target to conduct our experiment only on businesses having between 50-100 reviews, and sample for each category (e.g., hotels) the top 10 businesses with the highest number of reviews in the current filter.
The process finally forms a sample of 4966 reviews (31860 review sentences) supporting 50 Yelp businesses under 5 categories to be covered in our experiment.
We show additional statistics of our \textsc{Yelp} dataset in Table~\ref{table:yelp_statistics}

\begin{table}
    \centering
    \small
    \caption{Statistics of \textsc{Yelp}}
    \label{table:yelp_statistics}
    \begin{tabularx}{0.5\textwidth}{|
        l|
        >{\centering\arraybackslash}X|
        >{\centering\arraybackslash}X|
        >{\centering\arraybackslash}X|}
        \hline
            \textbf{Category} & \textbf{\# Reviews} & \textbf{\# Sentences} & \textbf{\# Sentences Per Review} \\ \hline
            Arts & 994 & 6000 & 6.04 \\ \hline
            Auto & 994 & 6196 & 6.23 \\ \hline
            Beauty & 995 & 6288 & 6.32 \\ \hline
            Hotels & 983 & 7145 & 7.27 \\ \hline
            Rest & 1000 & 6231 & 6.23 \\
        \hline
    \end{tabularx}
\end{table}

\section{Dimensions of KP Quality Evaluation}
\label{sec:kp_quality_dimensions}
This section provides detailed descriptions of tasks and dimensions involved in our manual evaluation of the KP textual quality.
Annotators were asked to perform a pairwise comparison between two sets of KPs, each taken from a different model, generated for a specific reviewed business entity considering a specific dimension.
The annotators must answer a comparative question with respect to the evaluating dimension. (e.g., \emph{Which of the two summaries captures better \dots{}}).
For each dimension, following~\citet{friedman-etal-2021-overview}, we calculate the ranking using the Bradley-Terry model~\citep{bradley_terry}, which predicts the probability of a given participant winning a paired comparison, based on previous paired comparison results of multiple participants, and thus allows ranking them.

\begin{itemize}
    \item \textsc{Validity}: The key point should be an understandable, well-written sentence representing an opinion of the users towards an aspect of the business entity. This would filter out sentences such as \emph{``It's rare these days to find that!''}.
    \item \textsc{Sentiment}: The key point should have a clear sentiment towards the business entity under reviewed. (either positive or negative). This would exclude sentences like \emph{``I came for a company event''}.
    \item \textsc{Informativeness}: It should discuss some aspects of the reviewed business and be general enough. Any key point that is too specific or only expresses sentiment cannot be considered a good candidate.
    Statements such as \emph{``Love this place''} or \emph{``We were very disappointed''}, which merely express an overall sentiment, should be discarded, as this information is already conveyed in the star rating. The KP should also be general enough to be relevant for other businesses in the domain. A common example of sentences that are too specific is mentioning the business name or a person's name (\emph{``Byron at the front desk is the best!''}).
    \item \textsc{SingleAspect}: It should not discuss multiple aspects (e.g., \emph{``Decent price, respectable portions, good flavor''}).
\end{itemize}

\begin{itemize}
    \item \textsc{Redundant}: Each KP should express a distinct aspect. In other words, there should be no overlap between the key points.
    \item \textsc{Coverage}: A set of KPs should cover a wide diversity of opinions relevant and representative of the reviewed business.
    \item \textsc{Faithfulness}: KPs should express reasonable and meaningful opinions to the reviewed business without hallucination. No conjecture or unfounded claims arise.
\end{itemize}

\section{Pairwise KP Quality Comparison Annotation Guidelines}
\label{sec:pairwise_kp_quality_annotation_guideline}
Below are the two summaries for a business in \emph{Arts \& Entertainment}, generated by two different summarization frameworks. Each summary contains several key points (i.e., salient points) generated summarizing the user opinions on different aspects. You are tasked to select which summary you think is better according to the below criteria.

\textbf{Business:} Saenger Theatre.

\textbf{Criteria:} REDUNDANCY. Each key point in the summary should express a distinct aspect. In other words, there should be no overlap between the key points.

\textbf{Summary A:} ['The Saenger Theater is a beautiful and stunning venue.', 'Comfortable seating.', 'Great shows.', 'Beautiful and impressive renovation.', 'Excellent acoustics and sound quality.', 'Technical issues during the performance.', 'Limited and uncomfortable bathroom space.', 'Show cancellations and disruptions.', 'Uncomfortable seats and high seat prices.', 'Disappointing theater experience.']

\textbf{Summary B:} ['The renovations of the theater were praised.', 'The theater had exceptional shows.', 'Canceled shows were criticized.', 'The venue is stunning.', 'The staff at the theater was great.', 'Limited space in the bathroom was criticized.', 'The setup of the bathrooms was odd.', "The theater's location received negative comments."]

The options are:
\begin{itemize}
\item Summary A
\item Summary B
\end{itemize}

\section{Key Point Matching Annotation Guidelines}
\label{sec:key_point_matching_annotation_guideline}
Below are the match annotation guidelines for (sentence, KP) pairs:\\

In this task you are presented with a business domain, a sentence taken from a review of a business in that domain and a key point.

You will be asked to answer the following question: does the key point match the sentence?

A key point matches a sentence if it captures the gist of the sentence, or is directly supported by a point made in the sentence.

The options are:
\begin{itemize}
\item Yes
\item No
\item Faulty key point (not a valid sentence or unclear)
\end{itemize}

\section{Comparative Analysis of KP Quality: A Visual Overview}
\label{sec:kp_quality_visual_overview}
\begin{figure*}[t!]
    \centering
    \subfloat[KPs as summaries for salient points from corpus]{%
    \includegraphics[width=1\textwidth]{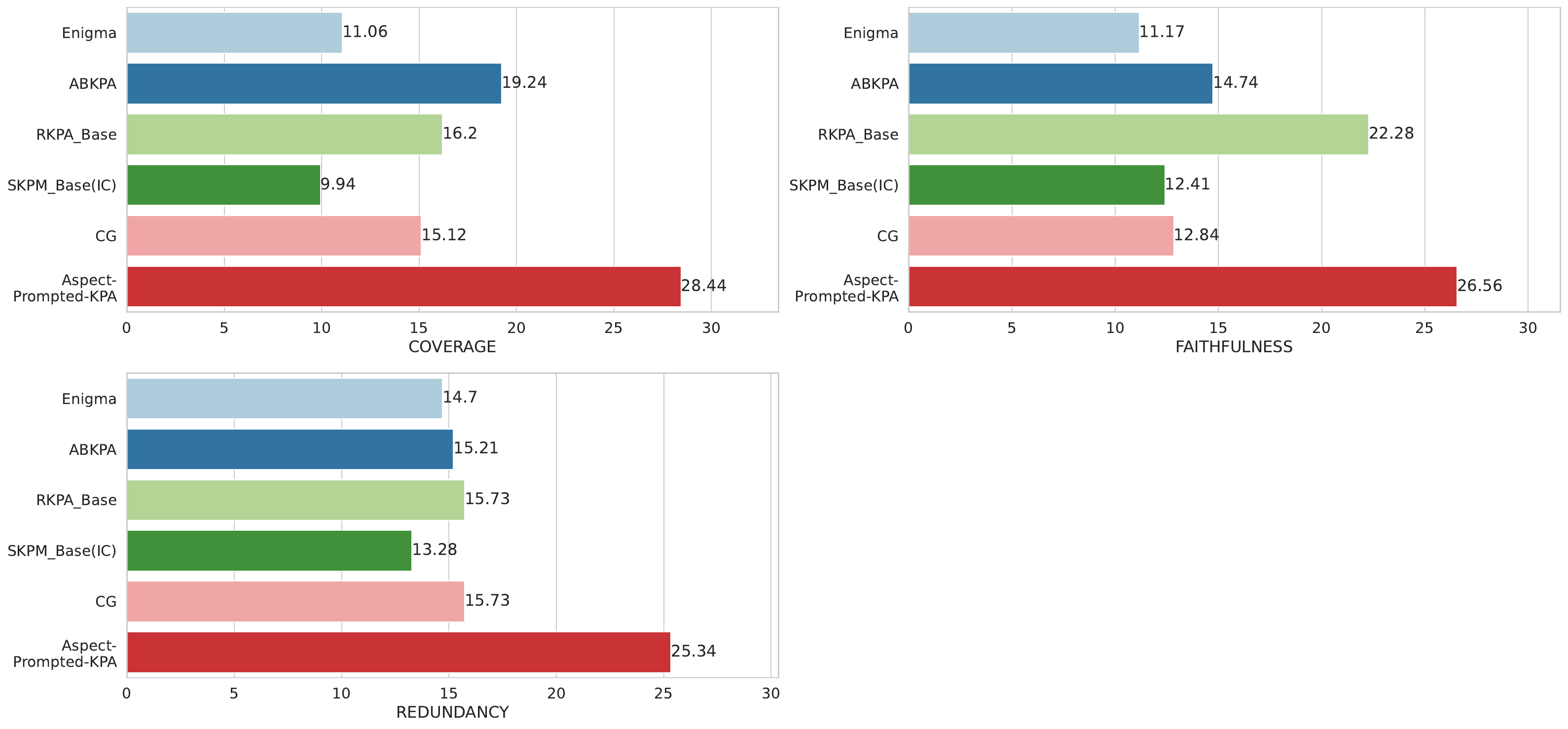}%
        \label{fig:a}%
        }%
    \vfill%
    \subfloat[KPs as summaries for reviews]{%
    \includegraphics[width=1\textwidth]{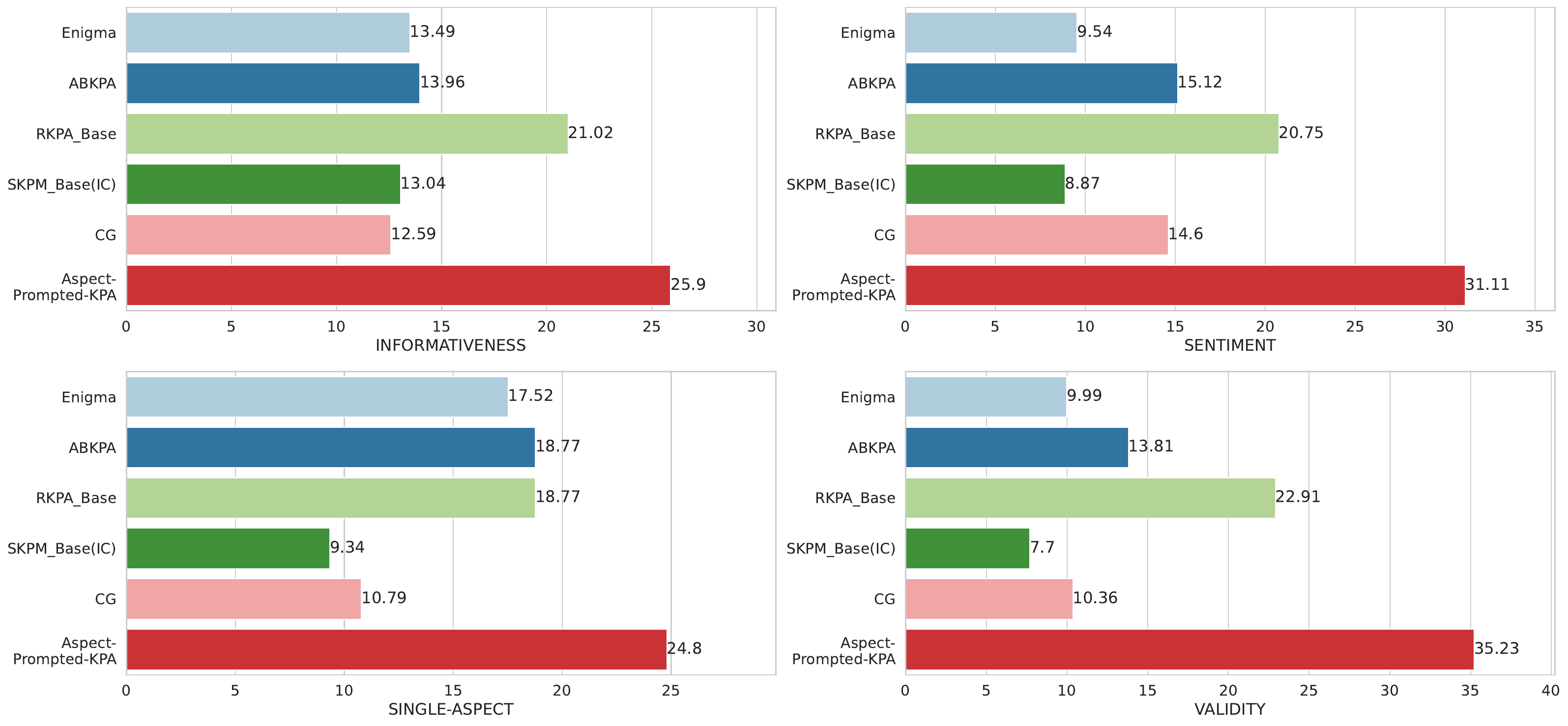}%
        \label{fig:b}%
        }%
    \caption{Bradley Terry scores of comparative human evaluation of different KPA frameworks on 7 dimensions in assessing how well they summarize the corpus (\ref{fig:a}) and provide KPs for reviews (\ref{fig:b}).
    }
    \label{fig:kp_quality_eval_results_fig}
\end{figure*}

Figure~\ref{fig:kp_quality_eval_results_fig} visualizes the Bradley Terry scores. as already presented in Table~\ref{table:kp_quality_eval_results_T}, in bar charts for more comprehensive view of our human evaluation results on different KPA systems.

\section{Summary of KPA Frameworks and Prompted Opinion Summarization Framework}
\label{sec:kp_summary_examples}
This section presents details of Table~\ref{table:top_negative_kps},
which shows some top 
negative KPs for all KPA systems, ranked by their prevalence and compares with the textual summary generated by the traditional prompted summarization framework (using \textsc{GPT3.5}) (CG).

\begin{table*}[h!]
    \centering
    \aboverulesep = 0pt
    \belowrulesep = 0pt
    \begin{tabularx}{\textwidth}{|X|X|X|X|X|}
    \hline
        \multicolumn{1}{|X}{\textbf{PAKPA}} & \multicolumn{1}{|X}{$\textbf{SKPM}_{\textbf{Base}}\textbf{(IC)+}$} & \multicolumn{1}{|X}{\textbf{Enigma+}} & \multicolumn{1}{|X}{\textbf{ABKPA}} & \multicolumn{1}{|X|}{\textbf{RKPA-Base}} \\ \hline
        Issues with the room and front desk service. & didn't work at all - the front desk staff was rude, rude, and!!! & They don't listen!!!! & Cons:* Very noisy rooms. &  Overall unprofessional and unorganized. \\ \hline
        Terrible hotel experience. & didn't have a receptionist at the front desk.!!! & Called front desk. &  Overall unprofessional and unorganized. & Carpet was stained and filthy. \\ \hline
        Difficult and expensive parking options. & a hotel is a "non smoking" hotel.!!! & They did not plan ahead! & And parking was also overpriced. & It didn't feel safe. \\ \hline
        Poor service and unresponsive staff. & I would never stay here again.!!! & Hotel is disgusting. & Poor hotel for the price. & are rude, slow and disrespectful. \\ \hline
        Issues with shower and bathroom cleanliness. & was a bit of a walk from the hotel to the parking lot.!!! & Would not recommend this hotel. & The food service was slow. & beds are very lumpy. \\ \hline
        \multicolumn{5}{|c|}{\dots} \\\hline
        \multicolumn{5}
        {|>{\hsize=\dimexpr5.5\hsize+2\tabcolsep+\arrayrulewidth\relax}X|}
        {
            \textbf{Recursive GPT-3-Chunking (CG):} \dots. However, negative aspects mentioned included \textbf{issues with room conditions}, \textbf{slow service}, \textbf{noise}, \textbf{safety concerns}, and \textbf{lack of amenities}. \dots
        }\\
        \hline
    \end{tabularx}
    \caption{\label{table:top_negative_kps}
    Top 5 negative-sentiment key points, produced by experimenting KPA systems, ranked by their prevalence on a ``Hotel'' business on \textsc{Yelp}, comparing with the textual summary created by the prompted opinion summarization framework (CG).
    }
\end{table*}
\begin{table*}[h!]
    \centering
    \aboverulesep = 0pt
    \belowrulesep = 0pt
    \begin{tabularx}{\textwidth}{|X|X|X|X|X|}
    \hline
        \multicolumn{1}{|X}{\textbf{PAKPA}} & \multicolumn{1}{|X}{$\textbf{SKPM}_{\textbf{Base}}\textbf{(IC)+}$} & \multicolumn{1}{|X}{\textbf{Enigma+}} & \multicolumn{1}{|X}{\textbf{ABKPA}} & \multicolumn{1}{|X|}{\textbf{RKPA-Base}} \\ \hline
        Excellent bakery with delicious treats. & has a good selection of pastries, pastries, pastries, and pastries!!! & Bread, baguettes, fresh. & Love love love this place. & Great baked sweets and breads. \\ \hline
        Delicious and diverse cake options. & has a good selection of pastries/cookies/cookies/c!!! & The best bread in Tucson. & Cappuccino and croissants are delish! & Prices are extremely reasonable! \\ \hline
        Friendly and efficient staff. & Sprouts' has a good selection of breads and pastries.!!! & You gotta go here!!! & Clean and well staffed. & They're worth the wait! \\ \hline
        Excellent prices. & Definitely recommend this place to anyone looking for a good!!! & The food is delicious. & Great baked sweets and breads. & Great food and flavor! \\ \hline
        Delicious baked goods. & I will definitely be back.!!! & Very friendly staff. & Prices are extremely reasonable! & Best friendly service, ever! \\ \hline
        Irresistible smells and incredible taste. & has the best bread in Tucson at a reasonable price.!!! & It was delicious! & Always hot and fresh tasting. & Familiar yet unique! \\ \hline
        Enchanting and beloved place. & I've been to this bakery for 20 years!!! & Nice old school bakery. & Great stop for lunch. & Amazing food and friendly service. \\ \hline
        \multicolumn{5}{|c|}{\dots} \\\hline
        \multicolumn{5}
        {|>{\hsize=\dimexpr5.5\hsize+2\tabcolsep+\arrayrulewidth\relax}X|}
        {
            \textbf{Recursive GPT-3-Chunking (CG):} \dots The bakery is highly regarded as the best in Tucson, with high-quality products. \dots Specific items like the baguette, sesame rolls, and dinner roll were highly rated for their taste, texture, and reasonable prices. \dots Customers appreciated the bakery\'s "old school" vibe, excellent prices, and consistently wonderful French bread and pastries. \dots Customers also praised the early opening hours, friendly staff, and variety of baked goods available. \dots
        }\\
        \hline
    \end{tabularx}
    \caption{\label{table:top_positive_kps}
    Top 7 positive-sentiment key points, produced by experimenting KPA systems, ranked by their prevalence on a ``Restaurant'' business on \textsc{Yelp}, comparing with the textual summary created by the prompted opinion summarization framework (CG). 
    }
\end{table*}

\end{document}